# MotionInput v2.0 supporting DirectX: A modular library of open source gesture-based machine learning and computer vision methods for interacting and controlling existing software with a webcam

Ashild Kummen[a], Guanlin Li[a], Ali Hassan[a], Teodora Ganeva[a], Qianying Lu[a], Robert Shaw[a], Chenuka Ratwatte[a], Yang Zou[a], Lu Han[a], Emil Almazov[a], Sheena Visram[b,c], Andrew Taylor[c], Neil J Sebire[c], Lee Stott[d], Yvonne Rogers[b], Graham Roberts[a], Dean Mohamedally[a]

[a]Department of Computer Science, University College London, London, United Kingdom

[b]UCL Interaction Centre, University College London, London, United Kingdom

[c]DRIVE Centre, Great Ormond Street Hospital for Children, London, United Kingdom

[d] Microsoft UK

## Abstract

**Introduction**. Touchless computer interaction has become an important consideration during the COVID-19 pandemic period. Despite progress in machine learning and computer vision that allows for advanced gesture recognition, an integrated collection of such open-source methods and a user-customisable approach to utilising them in a low-cost solution for touchless interaction in existing software is still missing. **Method**. In this paper, we introduce the MotionInput v2.0 application. This application utilises published open-source libraries and additional gesture definitions developed to take the video stream from a standard RGB webcam as input. It then maps human motion gestures to input operations for existing applications and games. The user can choose their own preferred way of interacting from a series of motion types, including single and bi modal hand gesturing, full-body repetitive or extremities-based exercises, head and facial movements, eye tracking, and combinations of the above. We also introduce a series of bespoke gesture recognition classifications as DirectInput triggers, including gestures for idle states, auto calibration, depth capture from a 2D RGB webcam stream and tracking of facial motions such as mouth motions, winking, and head direction with rotation. Three use case areas assisted the development of the modules: creativity software, office and clinical software, and gaming software. **Results**. A collection of open-source libraries has been integrated and provide a layer of modular gesture mapping on top of existing mouse and keyboard controls in Windows via DirectX and DirectInput. DirectX is a set of Application Programming Interface (APIs) available on Windows that give direct access to low-level hardware features including the controls of keyboard, mouse, and controller input.Four motion tracking modules have been packaged in this release; eye tracking, head movement tracking, hand gestures and full body exercises. The modules are presented to recognise a variety of human motions as input via a standard webcam, according to the preferences and needs of the user. Each gesture recognition triggers a unique input or series of inputs, optimised for existing DirectX enabled games and applications on Microsoft Windows. Currently, users can execute cursor commands: moving a cursor with their hands, eyes, or head direction, perform various clicking events with hand pinching or eye blinking and mouth opening gestures, and scroll with hand gestures or head movement. Physical detection of movement such as exercises that have extremities measured and repetitive behaviour are also measured as inputs for DirectX, for example jumping, squatting, cycling, air punches and kicks. Three additional key contributions arising from this work to be reported are: Compensating for hand jitters in air on a 2D plane, depth variation captured from a 2D webcam feed relayed as a user interface input, and an idle state for activating and deactivating gestures and features in software. For each of the three use case areas, specific variables for these open-source libraries are available in the common MotionInput graphical user interface for users to refine the sensitivity and set their relevant options. **Conclusion**. With ease of access to webcams integrated into most laptops and desktop computers, touchless computing be- comes more available with MotionInput v2.0, in a federated and locally processed method. This system is designed to work across a broad range of contexts with motion inputs; for example, in a hospital using complex user interfaces, playing games, exercising, for rehabilitation, and for creative tasks.

# 1 INTRODUCTION

We now interact with computer interfaces in our home environment for work, education, and gaming purposes, for prolonged periods and in new ways. To demonstrate the potential applications of recognition-based gestures we present University College London's (UCLs) MotionInput v2.0 supporting DirectX; a modular framework to describe four modules of gesture inputs for Windows based interactions. These use a regular webcam and open-source machine learning libraries to deliver low-latency input on existing DirectInput API calls within DirectX enabled games and applications on Microsoft Windows 10. This includes overriding mouse motions and keyboard presses. MotionInput supporting DirectX addresses some of the challenges of affordability and manual set up of existing hardware to do touchless gesture recognition. This paper presents the application architecture and explores MotionInput's capabilities with preliminary findings from the use of open-source machine learning solutions for detecting classifications of human motion activities.

# 2 METHOD

A literature review of the current state of the art in open-source gesture recognition technologies has been conducted. Several candidate libraries that have been investigated include MediaPipe[1], OpenVino, DLib and OpenCV[2], which have all been tested across gender and skin types. The libraries are federated in that they all process images locally on device and are not being processed on the cloud. Much of this machine learning work has been developed in Python. These Python based libraries have been brought into an application with a native front end designed in C# for Windows, compiled as a native executable with PyInstaller. The application developed has a settings control panel for switching state flags and key values in their respective libraries. Several gestures have been pre-designed and implemented in the Python code base. The taxonomy of these gestures has been identified under module headings such as exercise, head movement, eye-tracking, and hand gestures (Figure 1). The modules are connected by a common Windows based GUI configuration panel that exposes the parameters available to each module. This allows a user to set the gesture types and customise the responses and choose which modules to activate. These modalities mean that MotionInput reaches a broad user base across many use cases. For example, healthcare staff can navigate medical images touch free on a computer with mid-air hand gestures. Users with motor impairments would be able to use head movements or eye tracking to interact with their computer, and full-body rehabilitative exercises can be performed by playing games with skeletal tracking capabilities. Testing the different modules in their respective use cases has enabled us to build with preset values that can be configured by the user to their needs. Three use case areas have been examined: (1) creativity software (2) office and clinical software and (3) gaming software. Both standing and sitting modes of use have been developed in the controlling layer of the application.

# 3 PRELIMINARY RESULTS

## 3.1 Idle state and dynamic area of interest

This iteration of MotionInput is able to detect if a hand is in an idle or active state, by characterising an "idle hand" as that of being turned around (palm facing inward rather than towards the camera) or knuckles relaxed so that the tip of the index finger is resting below the tip of the thumb (or not present in the frame at all). This has been implemented to avoid the processing of unintended gestures. The system also auto calibrates based on distance to the camera, through the Dynamic Area of Interest (DAoI, see blue rectangle, Figure 2). The corners of the DAoI map to the corners of the screen and as you move further away the DAoI gets smaller to accommodate for a smaller area of use on screen.

---

*Email address:* D.Mohamedally@ucl.ac.uk (Dean Mohamedally)

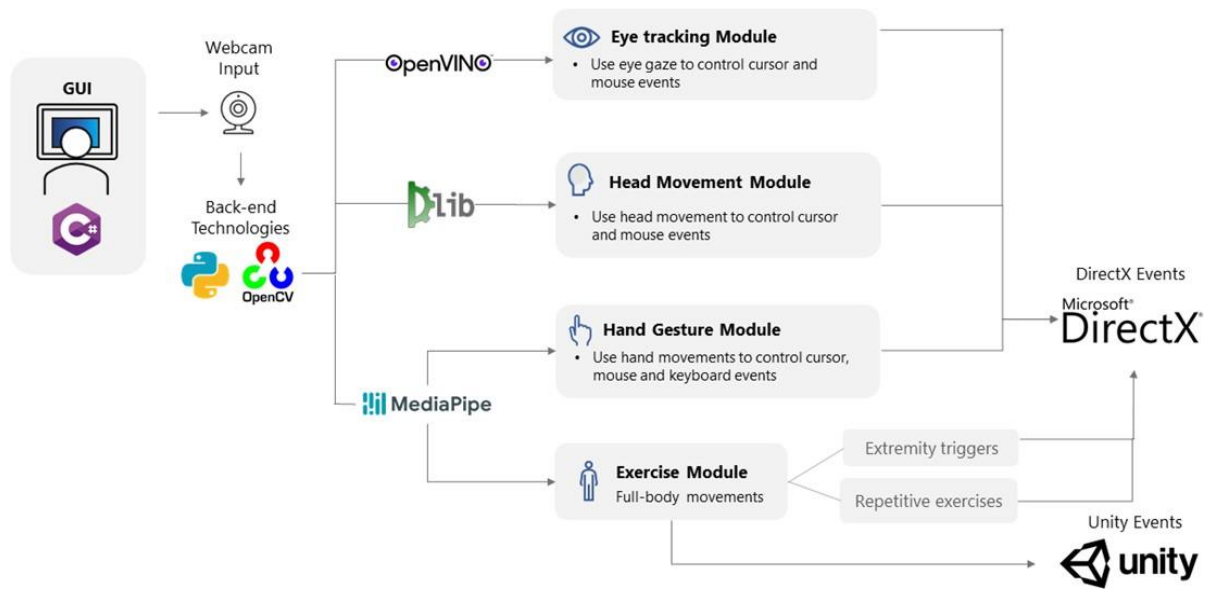

**Figure 1. System architecture of the prototype software**

### 3.2 Hand gesture module

The module has been developed with Mediapipe Hands[3], a convolutional neural network (CNN) that detects the presence of hands and landmarks on the hand (e.g., joints, fingertips, wrist). The model performs well across real-world settings when tested across genders and skin types. We use the hand landmarks obtained from Mediapipe Hands to classify hand gestures, such as detecting which joints are bent, which way the hand is facing, and distance between fingertips. By creating a set of gestures, the application can recognise, we can map gestures to specific keyboard commands or mouse events. For example, pinching your index finger and thumb together can trigger a click, allowing the user to interact with the computer without requiring a keyboard, mouse, or touch pad (Figure 2). The module will allow for gestures to be performed either sitting or standing up to 2 meters away from the camera, in line with the scope of MediaPipe's technological solution. Some of the limitations with using MediaPipe include that gesture recognition does not work as well in low-light conditions or with hands using gloves, but the solution remains an improvement to our previous iteration.

### 3.3 Exercise tracking module

The module permits users to design, customise and use their own movements, such as physical exercises that can be repetitious, to navigate their existing computer games. Repetitious exercises could include jumping, reaching out, squatting, but also cycling and rowing on various physical exercise equipment (Figure 2). A user can define what exercises would lead to a key press in software. This is done with a low latency and high accuracy. This module uses the MediaPipe Holistic library[4] to capture all the hand and pose landmarks to detect the positions of skeletal joints, and distances and angles between different sets of joints to classify gestures performed to execute mouse and keyboard events, such as the W, A, S and D direction keys in a game. For every frame of the webcam, the 3D coordinates of hand and pose landmarks are used to identify exercise gestures, confidence scores are calculated using deep machine learning algorithms to find the corresponding class of the gesture from the default and custom set of gestures. The time library also tracks time spent on the different classes of gesture to self-monitor progress to achieve an informative user experience of how much physical exercise has taken place.

### 3.4 Eye tracking module

We propose an auto-calibration method for eye-tracking that obtains the gaze estimation. Gaze estimation includes a face detection model, a face landmark detection model, a head position estimation model, and their relationships (Figure 2). Depth capture is then introduced, using 3D human pose estimation to capture the 3D position (x, y, z) of the user's nose tip. The gaze vector and depth information are then used to compute the location (x, y) on the screen (assuming the relative position between camera and screen is known). We anticipate that users will not need to calibrate each time; this is subject to accuracy of gaze estimation and depth capture.

### 3.5 Head tracking module

This module was built for users who are unable to use their hands to operate a mouse or track pad, in order to enable them to interact with a computer. Previous head tracking techniques use sensors attached to the user's head to measure head movement. However, thanks to advances in computer vision, we capture head movement without requiring wearable hardware. The head movement module is built using Dlib's prebuilt model5, which achieves rapid face-detection and accurately predicts 68 2D facial landmarks. Using these predicted landmarks of the face allows specific actions and facial expressions to be identified and mapped to specific mouse events. For example, by calculating the eye-aspect-ratio, a blink or a wink is detected, the mouth-aspect-ratio is similarly used to detect a yawn, and the left/right profile of a face to trigger mouse click events (Figure 2). In this module certain head movements of facial gestures are programmed as triggers to control the mouse cursor. Like the hand gesture module, limitations include low-light conditions which sometimes prevent facial landmark detection. On the other hand, facial landmarks, including the eyes, are perfectly detected when the user is wearing glasses. The face can be detected when sitting or standing in front of the camera at 2 meters as the maximum distance. Mapping facial expressions to keyboard events should be considered as further developments. Nevertheless, the development of the head movement module in v2 can be considered as a valuable tool for individuals whose hand mobility is restricted.

### 3.6 Free drawing and measurements

For accessibility, we can combine the use of hand gestures, head movements and eye-tracking, for example, to paint on a PC using common paint software. The aim is to allow users to use these features with any creative app, including those for art creation, and for music creation and direction. This is done by redirecting input via DirectInput in DirectX, to the mouse and keyboard events using MotionInput. This can also enable interaction with clinical applications in Radiology that use DICOM image viewers, where functionalities like those within painting software may be used as a measurement tool and a free drawing tool. Being able to move and click with high accuracy and low latency is of particular importance in this setting.

### 3.7 Depth capture from a 2D image signal

Depth capture in MotionInput acquires depth information from camera coordinates for the distance from the RGB camera to user. For hand gestures, depth information enables recognition of the actual position of each joint to better capture the gesture. Eye tracking and head tracking use the actual coordinate from depth information to compute the position on the screen that the user is looking.

For hand gestures we measure the size of the palm and use that to infer the distance from the hand to the camera. This is used to auto calibrate the area in which the user can gesture to move the cursor within (termed Dynamic Area of Interest). For eye tracking, we propose two methods for depth capture. The first is capturing depth from a user's iris, which is inspired by the project Depth-from-Iris from Mediapipe[1]. It can estimate the distance from camera to Iris with less than 10% error. The principle is that a human horizontal iris diameter is consistent across users, removing a need to calibrate to each user and Depth-from-Iris measures the depth of the user's eyes. The second method uses 3D human pose estimation to capture the depth of users. This method is inspired by a GitHub project Real-time 3D Multi-person Pose Estimation, which can estimate the coordinates of 21 key points in a human, including two eyes and nose. The coordinates of two eyes with respect to the camera are necessary for eye tracking. For this project's purpose, two neural networks can be used to capture depth for eye-tracking. One is based on Depth-from-Iris, while the other is based on 3D human pose estimation.

## 4 CONCLUSION

In this paper, we have presented MotionInput v2.0 supporting DirectX, as a standard set of interaction gestures which can be used to navigate, control and make use of existing software and games on Windows PCs. The application has classified four gesture module types, based on, and configured with three open-source Python-based gesture libraries. All machine learning and computer vision with these libraries are federated and processed on a local device and the webcam feed is processed and discarded; the solution does not connect to any cloud provider for processing meaning there are less data safety risks for use in clinical and healthcare related settings.

We have also presented a series of methodologies for gesture detection. MotionInput v2.0 enables accessibility for touchless human-computer interactions that at the time of publication are currently not available in a coherent packaged form for end users. The ability to use mainstream low-cost webcams for motion input, combined with these open-source machine learning libraries for tracking user motions and then relay that as input into existing software as though it was connected as mouse and keyboard input, has the potential to change the way that users interact with computers systems.

The project website is available at www.motioninput.com

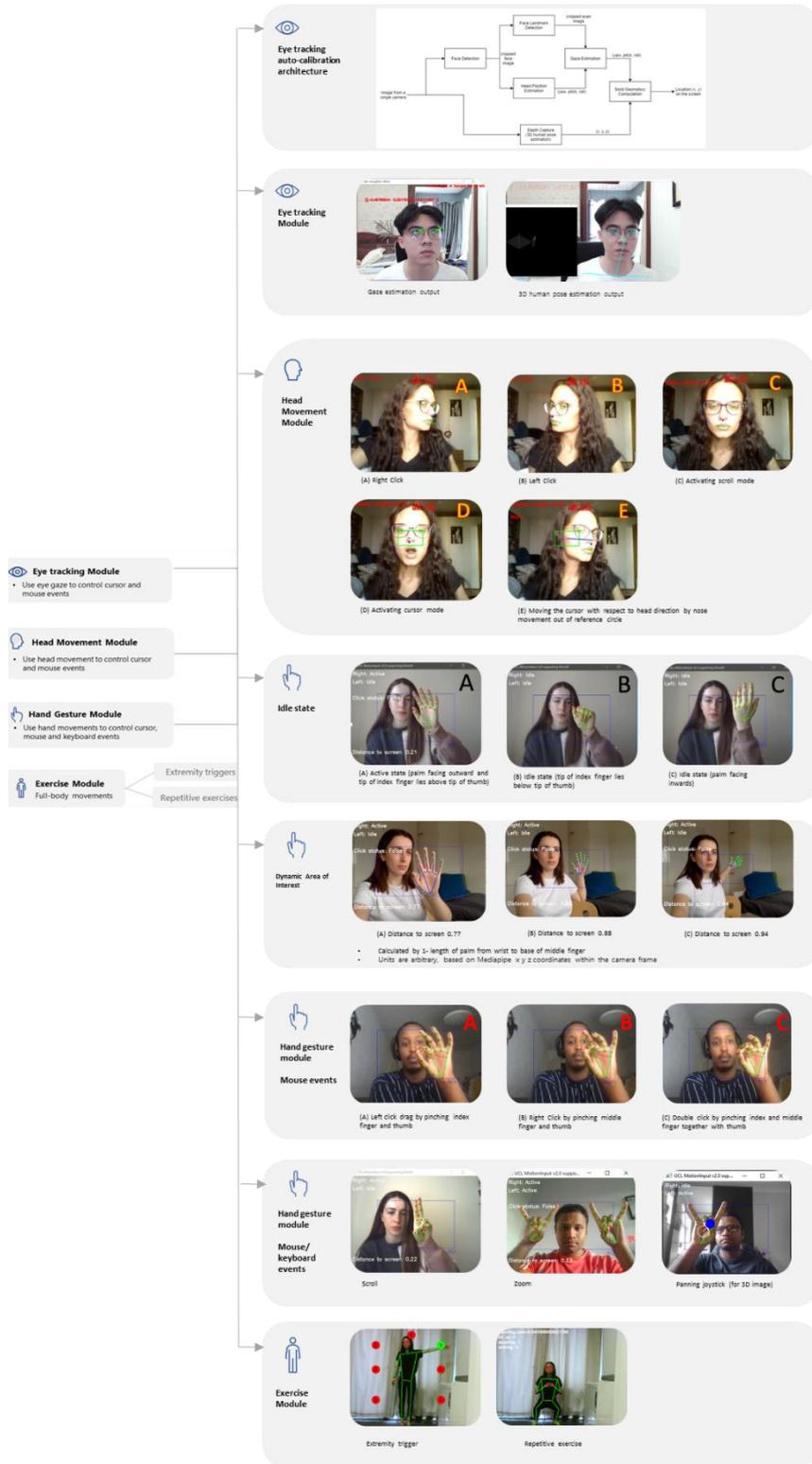

**Figure 2. MotionInput v2.0 gestures; illustrated by module: eye tracking, head movement, hand gestures, exercise.**